\documentclass{article}

\usepackage{arxiv}

\usepackage[utf8]{inputenc} 
\usepackage[T1]{fontenc}    
\usepackage{hyperref}       
\usepackage{url}            
\usepackage{booktabs}       
\usepackage{amsfonts,amsmath}       
\usepackage{nicefrac}       
\usepackage{microtype}      
\usepackage{lipsum}
\usepackage{graphicx}
\usepackage{bm}
\graphicspath{ {./images/} }

\title{Physics-informed Gaussian process model for Euler-Bernoulli beam elements}

\author{
 Gledson Rodrigo Tondo$^{\star}$ \\
  Bauhaus-Universität Weimar\\
  Weimar, Germany \\
   \And
 Sebastian Rau \\
  Bauhaus-Universität Weimar\\
  Weimar, Germany \\
  \And
 Igor Kavrakov \\
  Bauhaus-Universität Weimar\\
  Weimar, Germany \\
  \And
 Guido Morgenthal \\
  Bauhaus-Universität Weimar\\
  Weimar, Germany \\
}

\begin{document}
\maketitle
\begin{abstract}
A physics-informed machine learning model, in the form of a multi-output Gaussian process, is formulated using the Euler-Bernoulli beam equation. Given appropriate datasets, the model can be used to regress the analytical value of the structure’s bending stiffness, interpolate responses, and make probabilistic inferences on latent physical quantities. The developed model is applied on a numerically simulated cantilever beam, where the regressed bending stiffness is evaluated and the influence measurement noise on the prediction quality is investigated. Further, the regressed probabilistic stiffness distribution is used in a structural health monitoring context, where the Mahalanobis distance is employed to reason about the possible location and extent of damage in the structural system. To validate the developed framework, an experiment is conducted and measured heterogeneous datasets are used to update the assumed analytical structural model.
\end{abstract}


\section{Introduction}
Machine learning has been extensively applied in structural engineering, especially in the structural health monitoring (SHM) field, as the availability of data collected from sensors increases~\cite{worden2007application}. The heterogeneity of the collected datasets along with the knowledge of physical relations between them, usually represented as partial differential equations (PDEs), have motivated the recent use of physics-informed machine learning models to extract meaningful information from measured data. In these algorithms, the governing PDE is built into the machine learning model, effectively integrating measurements and mathematical models~\cite{karniadakis2021physics}. Gaussian processes (GP)~\cite{williams2006gaussian} have been extensively used for such a task, as they offer a non-parametric probabilistic view of the modelling scheme, which is usually framed in a Bayesian manner~\cite{raissi2017inferring,raissi2018numerical}. Particular applications of the physics-informed GPs have been observed as an analytical model of the latent curvature in a sleeper beam and the inverse problem of identifying the Reynolds number of a CFD simulation~\cite{gregory2019synthesis,raissi2018hidden}. In this paper, a physics-informed Gaussian process model for the Euler-Bernoulli beam formulation is developed to simultaneously infer physical quantities of interest, whilst considering the problem of identifying the correct structural bending stiffness. The model is defined in section 2, along with the optimization strategies, while section 3 is used to demonstrate the inference capabilities and tolerances to measurement noise in a numerical and controlled manner. Section 4 utilizes experimental measurements to update an analytical model of a steel beam.

\section{Physics-informed GP model of an Euler-Bernoulli beam}

\subsection{Model definition}

Consider the general linear form of the Euler-Bernoulli beam equation:

\begin{equation}
    EI \frac{d^4 u(\bm{x})}{d\bm{x}^4} = q(\bm{x}),
\end{equation}

where $u(\bm{x})$ are the beam deflections at positions $\bm{x}$ due to a given input load $q(\bm{x})$ and $EI$ is the structural bending stiffness, corresponding to Young’s modulus of elasticity $E$ and the second moment of area $I$.

In the physics-informed model derived here, the deflections $u(\bm{x})$ are represented as a zero-mean Gaussian process:

\begin{equation}
    u(\bm{x})  \sim \mathcal{GP}(\bm{0},k_{uu} (\bm{x},\bm{x}';\bm{\theta})),  
\end{equation}

where $k_{uu}$ is a covariance kernel parametrized by the values in $\bm{\theta}$. Several kernels are available in the literature, and without loss of generality, the deflection model is here defined with the universal squared exponential kernel

\begin{equation}
    k_{uu} (\bm{x},\bm{x}';\bm{\theta})) = \sigma_s^2 \mathrm{exp} \left( - \frac{1}{2} \left( \frac{\bm{x} - \bm{x}'}{\ell} \right) \right),
\end{equation}

where $\theta = \lbrace \sigma_s, \ell \rbrace$ contains the kernel standard deviation $\sigma_s$ and the length scale $\ell$, which controls the covariance smoothness~\cite{williams2006gaussian}.

By exploiting the linear aspect of GPs, a similar model can be derived for the forces:

\begin{equation}
    q(\bm{x})  \sim \mathcal{GP}(\bm{0},k_{qq} (\bm{x},\bm{x}';\bm{\theta}, EI)),  
\end{equation}

where the kernel $k_{qq}$ is derived from $k_uu$ through the beam model differential equation, such that 

\begin{equation}
    k_{qq} (\bm{x}, \bm{x}'; \bm{\theta}, EI) = EI \frac{d^4}{d \bm{x}^4} \left( EI \frac{d^4}{d \bm{x}'^4} k_{uu} (\bm{x}, \bm{x}'; \bm{\theta}) \right). 
\end{equation}

The cross-covariance functions between the applied loads and the resulting deflections can be derived as

\begin{equation}
    k_{uq} (\bm{x}, \bm{x}'; \bm{\theta}, EI) = EI \frac{d^4}{d \bm{x}'^4} k_{uu} (\bm{x}, \bm{x}'; \bm{\theta}) . 
\end{equation}

In a similar manner, the Gaussian process models for other related quantities of interest provided by the model, such as rotations $r$, strains $\epsilon$, bending moments $m$ and shear forces $v$ can be derived from $k_{uu}$ through the application of the respective linear differential operator. The complete Euler-Bernoulli beam model is then given by the multi-output GP

\begin{equation}
    \left[ \bm{u}, \bm{r}, \bm{\epsilon}, \bm{m}, \bm{v}, \bm{q} \right]^T = \mathcal{GP} (\bm{\mu}_p, \bm{K}_p) , 
\end{equation}

with the prior mean $\mu_p= \left[ \bm{0},\bm{0},\bm{0},\bm{0},\bm{0},\bm{0} \right]^T$ and the prior covariance matrix $\bm{K}_p$ formed as

\begin{equation}
		\label{eq:allKernelsCombined}
		\bm{K}_p = 
		\begin{bmatrix}
			\bm{K}_{ww}^{n} \! & \bm{K}_{w r} \! & \bm{K}_{w \epsilon} \! & \bm{K}_{w \!m} \! & \bm{K}_{wv} \! & \bm{K}_{wq} \!\\
			\bm{K}_{r w} \! & \bm{K}_{rr}^{n} \! & \bm{K}_{r \epsilon} \! & \bm{K}_{r \! m} \! & \bm{K}_{r v} \! & \bm{K}_{r q} \!\\
			\bm{K}_{\epsilon w} \! & \bm{K}_{\epsilon r} \! & \bm{K}_{\epsilon\epsilon}^{n} \! & \bm{K}_{\epsilon \! m} \! & \bm{K}_{\epsilon v} \! & \bm{K}_{\epsilon q} \!\\
			\bm{K}_{m \! w} \! & \bm{K}_{m \! r} \! & \bm{K}_{m \! \epsilon} \! & \bm{K}_{m \!m}^{n} \! & \bm{K}_{m \! v} \! & \bm{K}_{m \!q} \!\\
			\bm{K}_{vw} \! & \bm{K}_{v r} \! & \bm{K}_{v \epsilon} \! & \bm{K}_{v \! m } \! & \bm{K}_{vv}^{n} \! & \bm{K}_{vq} \!\\
			\bm{K}_{qw} \! & \bm{K}_{q r} \! & \bm{K}_{q \epsilon} \! & \bm{K}_{q \! m} \! & \bm{K}_{qv} \! & \bm{K}_{qq}^{n} \!
		\end{bmatrix}.
	\end{equation}

where $K_{ab}= k_{ab} (\bm{x},\bm{x}';\bm{\theta},EI)$ is a covariance matrix generated with its respective kernel function. To account for measurement noise in each dataset, an additional diagonal block matrix $k_a^n$ is added to the covariance of each kernel matrix, calculated as:

\begin{equation}
    \bm{K}_a^n = \sigma_a^2 \delta_{ij} (\bm{x}, \bm{x}'), 
\end{equation}

where $\delta_{ij}$ is the Kronecker delta operator and $\sigma_a$ is the noise standard deviation value for dataset a, such that $\bm{K}_{aa}^n=\bm{K}_{aa}+\bm{K}_{a}^n$. The standard deviation for each dataset, when unknown, is included in a parameter vector $\bm{\psi}$ and becomes an optimizable variable within the model.

Accounting for boundary conditions (BCs) in the model is possible through the application of a modified Green’s function to the kernel equation~\cite{raket2021differential,sarkka2011linear}, which restricts the generated model to a single structural system. Instead, in this work BCs are accounted for through the creation of an artificial dataset, at the appropriate locations, that measures the BC values in a noise-less manner.

\subsection{Model selection}

The selection of an appropriate model in face of the measured noisy data involves the optimization of the parameter vector $\bm{\psi}={\bm{\theta},EI,\bm{\psi}}$, that is, the kernel values, the bending stiffness and the noise in each dataset, respectively. For that matter, a probability distribution $p(\bm{\psi})$ can be created to represent the prior beliefs on each of the model parameters. Following the assumption of independence in the parameters $\bm{\psi}$, the log form of this distribution is calculated as:

\begin{equation}
    \mathrm{log} p(\bm{\psi}) = \sum_i \mathrm{log} p(\psi_i).
\end{equation}

Moreover, when data is presented to the model in the form of $\bm{y}=[\bm{u},\bm{r},\bm{\epsilon},\bm{m},\bm{v},\bm{q}]^T$, measured at locations $\bm{x}=[\bm{x}_u, \bm{x}_r, \bm{x}_{\epsilon}, \bm{x}_m, \bm{x}_v, \bm{x}_q]^T$, for $\bm{x},\bm{y} \in \mathcal{R}^{N \times 1}$, the log-likelihood can be analytically calculated by:

\begin{equation}
    \mathrm{log} p(\bm{y} | \bm{x}, \bm{\psi}) = -\frac{1}{2} \bm{y}^T \bm{K}_p^{-1} \bm{y} - \frac{1}{2} \mathrm{log} |\bm{K}_p| - \frac{N}{2} \mathrm{log} 2 \pi,
\end{equation}

where $\bm{K}_p$ is the global covariance matrix as shown in equation 8, calculated for all datasets contained in $\bm{x}$ and $\bm{y}$~\cite{williams2006gaussian}.

Following Bayes’ rule, the log-prior and log-likelihood can be combined to generate a posterior distribution that reflects the influence of measured data in the model parameters, such that

\begin{equation}
    \mathrm{log} (\bm{\psi} | \bm{y}, \bm{x}) \propto \mathrm{log} \ p(\bm{y} | \bm{\psi}, \bm{x}) + \mathrm{log} \ p (\bm{\psi}).
\end{equation}

The formulation above is proportional to the true log-posterior up to a value defined by the marginal likelihood of the system, which is constant with respect to the parameters $\bm{\psi}$. The marginal likelihood takes the form of an intractable integral for most probabilistic systems, and therefore the parameter selection is carried out as an optimization problem given by:

\begin{equation}
    \bm{\psi} = \mathrm{argmax}_{\bm{\psi}} \ \mathrm{log} \ p(\bm{\psi} | \bm{y}, \bm{x}),
\end{equation}

which is popularly known as maximum a posteriori estimation. The maximization of this posterior distribution is estimated numerically in a probabilistic approach through the Metropolis-Hastings (MH) algorithm~\cite{hastings1970monte}. The algorithm is a Markov chain Monte Carlo method that returns a sequence of auto-correlated random samples of the parameters $\bm{\Psi}=[\bm{\psi}_0,\bm{\psi}_1,...,\bm{\psi}_k ]^T$, which are representative of the true posterior distribution $p(\bm{\psi}|\bm{y},\bm{x})$.

\subsection{Model inference}

Predictions of quantities $\bm{y}_{\star}$ at unobserved locations $\bm{x}_{\star}$ can be made by conditioning the predictive distribution on the noisy observations used during training, and integrating over the identified parameter distribution,

\begin{equation}
    p(\bm{y}_{\star} | \bm{x}_{\star}, \bm{y, \bm{x}}) = \int p(\bm{y}_{\star} | \bm{x}_{\star}, \bm{y}, \bm{x}, \bm{\psi}) p(\bm{\psi} | \bm{y}, \bm{x}) d \bm{\psi}.
\end{equation}

The inner probability distribution in the above formulation corresponds to the standard GP predictive posterior when no uncertainty in the parameters $\bm{\psi}$ exists. Due to the Gaussianity assumptions of the GP model, this probability takes the closed form of $p(\bm{y}_{\star} | \bm{x}_{\star}, \bm{y}, \bm{x}, \bm{\psi}) = \mathcal{N}(\bm{\mu}_{\bm{y}^{\star}},\bm{K}_{\bm{y}^{\star}})$, with the mean $\bm{\mu}_{\bm{y}^{\star}}$ and the covariance matrix $\bm{K}_{\bm{y}^{\star}}$ calculated by, respectively,

\begin{equation}
    \bm{\mu}_{\bm{y}^{\star}} = \bm{K}_{\star}^T \bm{K}_p^{-1} \bm{y},
\end{equation}

\begin{equation}
    \bm{K}_{\bm{y}^{\star}}= \bm{K}_{\star \star}-\bm{K}_{\star}^T \bm{K}_p^{-1} \bm{K}_{\star},
\end{equation}

where $\bm{K}_{\star}=[k_{au} (\bm{x}_{\star},\bm{x}'),...,k_{aq} (\bm{x}_{\star},\bm{x}')]^T$ is the cross-covariance matrix between unobserved locations and training positions for all training datasets, and $\bm{K}_({\star}{\star})=k_{aa} (\bm{x}_{\star},\bm{x}'_{\star})$ is the self-covariance matrix, for the particular quantity of interest $a$~\cite{williams2006gaussian,lalchand2020approximate}. 
A closed-form solution of Equation 14 is generally intractable, and therefore the mean and covariance parameters of the predictive distribution are estimated as:

\begin{equation}
     p(\bm{y}_{\star} | \bm{x}_{\star}, \bm{y}, \bm{x}) \approx \frac{1}{N} \sum_i^N p(\bm{y}_{\star} | \bm{x}_{\star}, \bm{y}, \bm{x}, \bm{\psi}_i),
\end{equation}

where $\bm{\psi}_i \sim p(\bm{\psi} | \bm{y}, \bm{x})$ are draws from the parameter posterior approximated with the MH algorithm~\cite{lalchand2020approximate}. The predictive posterior finally takes the form of a multivariate mixture of Gaussians, following the assumption of Gaussian noise in the dataset $\bm{y}$.

\section{Numerical experiments}

\begin{figure}[h]
  \centering
  \includegraphics{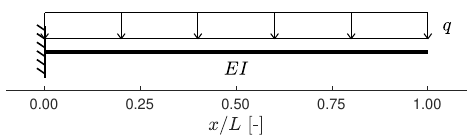}
  \caption{Benchmark structure for a numerical case: a cantilever beam with constant stiffness $EI$ and length $L$, subjected to a uniformly distributed load of magnitude $q$.}
  \label{fig:fig1}
\end{figure}

In this section, a numerical investigation of the presented method is carried out for a cantilever beam of length L with constant bending stiffness EI, subjected to a uniformly distributed load of magnitude q, as shown in Figure~\ref{fig:fig1}. For such a case an analytical solution exists, such that

\begin{equation}
    u(x)= \frac{qx^2}{24EI} x^2-4Lx+6L^2 .
\end{equation}

This solution is further used as a benchmark for the model’s predictions, along with the true bending stiffness EI, which is compared to the value regressed by the GP model. The software Matlab is used for the numerical implementation of the method.

\subsection{Stiffness regression and latent function inference}

In this work, prior knowledge on the noise values and the kernel parameters is modelled with a uniform distribution $p(\bm{\psi}_i )= \mathcal{U}(-\infty,\infty)$, while the bending stiffness is treated in constrained manner as $p(\psi_EI )= \mathcal{U}(0.1 EI_\mathrm{true},2.0 EI_\mathrm{true})$. It is further assumed that four deflection sensors, equally spaced throughout the length of the structure, monitor the system. These sensors normally operate under the influence of environmental conditions and, due to the nature of the measurement, present a variable level of uncertainty in their outputs. Noise values are herein simulated assuming a Gaussian model $u_n~\mathcal{N}(u_\mathrm{ana} (x),\sigma_n^u)$, where $u_\mathrm{ana} (x)$ is the analytical solution and $\sigma_n^u$ is a standard deviation given by

\begin{equation}
    \sigma_u^n = \frac{|\hat{u}|}{\mathrm{SNR}}
\end{equation}

where $|\hat{u}|$ is the maximum absolute displacement at the tip of the cantilever beam, and SNR is a signal-to-noise ratio parameter. 

\begin{figure}[h]
  \centering
  \includegraphics{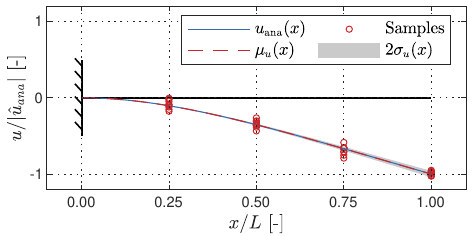}
  \caption{Normalized mean $\mu_u$ and standard deviation $\sigma_u$ of the displacement field predictions based on 5 data points collected by 4 noisy sensors along the structure.}
  \label{fig:fig3}
\end{figure}

By providing the load value q, along with a set of sensor readings obtained from different sensor locations along the structure, and contaminating the measurements with a noise defined by a SNR=10, the GP model can be trained and used for further inference. A sample of the predicted displacement field is shown in Figure~\ref{fig:fig3}. Despite the noisy measurements, the GP model is able to accurately predict the displacement values along the length of the beam. In addition, the model uncertainty, represented by the prediction’s standard deviation $\sigma_u$, increases along the length of the structure. This effect reflects the boundary condition assumption, introduced as an additional artificial noise-less sensor located at $x/L=0$.

\begin{figure}[h]
  \centering
  \includegraphics{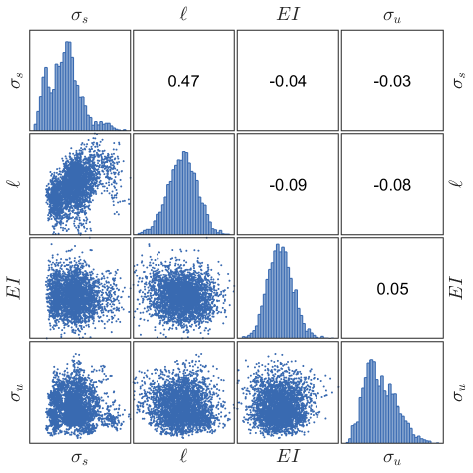}
  \caption{Correlation matrix of the probabilistic model parameters optimized using the MH algorithm. The kernel values $\sigma_s$ and $\ell$ show a positive degree of correlation, whereas the remaining parameters are uncorrelated.}
  \label{fig:fig9}
\end{figure}

Once the model is optimized, a set of auto-correlated parameters is returned from the MH algorithm, which approximates the posterior parameter distribution $p(\bm{\psi}|\bm{y},\bm{x})$. The results, shown in Figure~\ref{fig:fig9}, reflect the GP model uncertainty on each of the parameters in question.

\begin{figure}[h]
  \centering
  \includegraphics{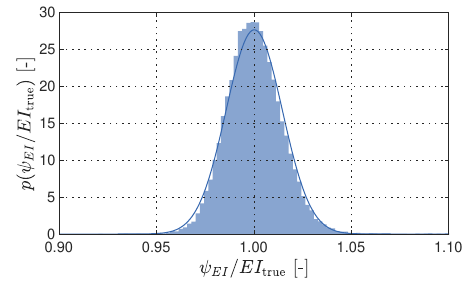}
  \caption{The regressed probabilistic stiffness value of the structure. The distribution's mean value approaches the correct stiffness $EI_\mathrm{true}$, while the standard deviation represents the model uncertainty.}
  \label{fig:fig4}
\end{figure}

Of particular interest in the parameter model is the stiffness distribution, shown in Figure~\ref{fig:fig4}. The mean value of $p(\psi_EI)$ approximates the true stiffness $EI_{\mathrm{true}}$ with a 0.63\% error. The standard deviation, calculated as $0.023EI_\mathrm{true}$, reflects the model uncertainty on the stiffness and is a function of the training dataset. Further, a normalized version of the stiffness distribution is defined as$ p(\psi_{EI}/EI_\mathrm{true}) = \mathcal{N}(\mu_{EI},\sigma_{EI})$, and the Mahalanobis distance, given by

\begin{equation}
    d_M = \sqrt{\frac{(\mu_{EI}-1)^2}{\sigma_{EI}^2}},
\end{equation}

is used to evaluate the quality of the predicted stiffness distribution. This formulation doesn’t only reflect the accuracy of the mean value but also penalizes uncertain stiffness models. The Mahalanobis distance of the stiffness distribution in Figure~\ref{fig:fig4} was calculated as $d_M=0.40$.

\begin{figure}[h]
  \centering
  \includegraphics{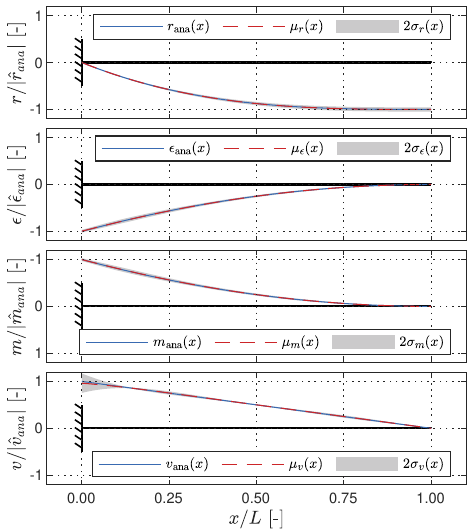}
  \caption{Mean $\mu$ and standard deviation $\sigma$ of the inferences on the latent rotations $r$, strains $\epsilon$, moments $m$ and shear forces $v$, normalized by the maximum absolute analytical result of the respective physical quantity.}
  \label{fig:fig8}
\end{figure}

In addition to the regressed parameters, the fully trained GP model can also be used to infer physical quantities that were not directly measured, as shown in Figure~\ref{fig:fig8}. 

Similarly to the displacement case, the mean values predicted for rotations, strains, bending moments and shear forces are in accordance with the analytical results, with a root mean squared error (RMSE) calculated for each of the normalized predictions in the order of $10^(-3)$. The standard deviations of the rotation predictions follow the trend of the displacement values, increasing along the length of the structure, as a result of the boundary condition enforced as $r(x=0)=0$~rad. Similarly, the strains, bending moments and shear values show an increase in uncertainty closer to the support location, once again, due to the boundary conditions $m(x=L)=0$~Nm and $v(x=L)=0$~N.

\subsection{Influence of noise and number of data points}

The training of this physics-informed GP model takes place iteratively, by consecutive evaluations of the probabilistic a posteriori distribution in the Markov chain of the Metropolis-Hastings algorithm. A drawback of a standard Gaussian process model is that, for every model evaluation, a computational cost of $\mathcal{O}(N^3)$ is involved. 

\begin{figure}[h]
  \centering
  \includegraphics{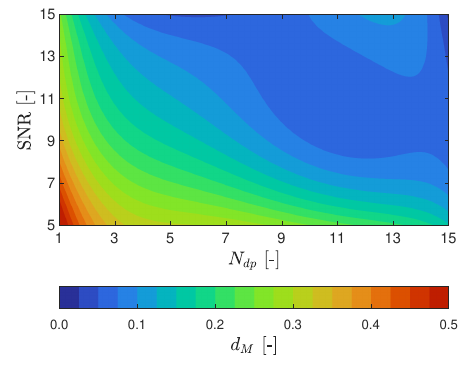}
  \caption{Influence of measurement noise and the number of data points provided to the algorithm. A total of 4 equidistant sensors were used in this study, as shown in Figure~\ref{fig:fig3}}
  \label{fig:fig5}
\end{figure}

Considering typical sampling rates for displacement sensors, a large amount of data can be quickly generated, and the usage of the GP model becomes unfeasible. Therefore, it is important to reduce the size of the dataset provided to the GP algorithm during training, while retaining a good prediction accuracy. An obvious correlation exists between the number of provided data points $N_{dp}$ and the level of noise of the measurements, which is numerically controlled by the signal-to-noise ratio (SNR). In Figure~\ref{fig:fig5} the interaction between noise and data size is observed. Considering four equidistant sensors (cf. Figure~\ref{fig:fig3}), the GP model has a clear improvement in accuracy when higher quality data, and more data points, are used as inputs. The trade-off exists as a model response to correctly identify the mean and standard deviation values, given the noisy points provided.

\subsection{Damage identification capabilities}

\begin{figure}[h]
  \centering
  \includegraphics{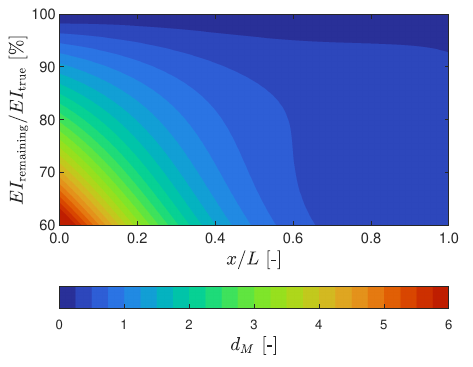}
  \caption{Damage study: influence of stiffness reduction and corresponding location along the structure on the model's predictions.}
  \label{fig:fig7}
\end{figure}

The parameter study presented in section 3.2 allows for an informed decision on sensor quality, as well as the computational costs of the GP model, given a minimum desired stiffness prediction accuracy. If a good prediction is guaranteed, and continuous monitoring of a structure is available, then the generated model can be used to detect, locate and estimate the severity of eventual damages.
Damage is estimated through the deviation of the original numerical stiffness $EI_{\mathrm{true}}$ using the Mahalanobis distance, calculated as per Equation 20. To simulate damage, a finite element model of the cantilever beam is generated, with a total of 20 elements. The damage is simulated as a reduction in the bending stiffness of up to 40\% of the original structure, in one element at a time, for all the finite elements in the model. In Figure~\ref{fig:fig7} the deviations of the original structural stiffness are shown, in terms of $d_M$, as a function of damage location and extent. 

The results indicate an increase in the Mahalanobis distance value for progressive reductions of remaining stiffness and damage locations closer to the support. Given that the structural system is a cantilever, the system’s displacement response is more sensitive to changes in close to the support. For damage cases at locations $x>0.60L$, the indication of damage may be obscured by the model’s prediction error, as a function of the noise levels. Nevertheless, given a damage condition closer to the support, the Mahalanobis distance allows for an informative decision of the location of the damage. Once the location is known, the damage extent identification can be achieved, given the linearity condition of the system.

\section{Experimental model updating}

\begin{figure}[h]
  \centering
  \includegraphics[width=\textwidth]{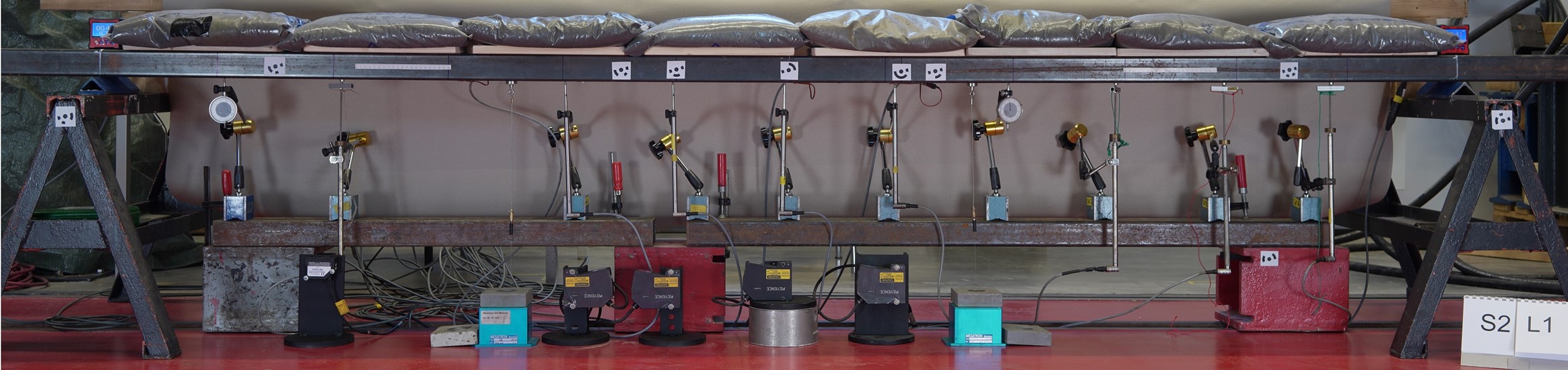}
  \caption{Experimental set-up: deflection, inclination and strain sensors are installed in a simply-supported steel beam. A static load uniformly distributed is applied with bags of steel spheres, resting on wooden plates.}
  \label{fig:fig10}
\end{figure}

To validate the proposed model, an experimental study conducted on a simply supported steel beam is now presented. The true bending stiffness is initially unknown but estimated through material and geometrical parameters to be $EI_\mathrm{init}$. The goal is to use measurement data and provide an optimized version of the bending stiffness value, in a model updating sense. Figure~\ref{fig:fig10} displays the structure and measurement setup. Different deflection sensors, such as laser deflectometers, displacement transducers, draw wire transducers and dial gauges were used to monitor the deflections due to the applied load. 

\begin{figure}[h]
  \centering
  \includegraphics{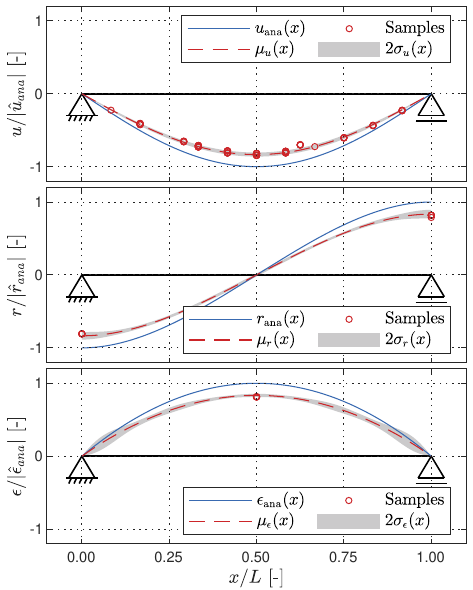}
  \caption{Normalized inference on the trained model using the experimental results, as compared to the initial, analytical model using $EI_\mathrm{init}$, shown in blue.}
  \label{fig:fig12}
\end{figure}

The degree of noise that contaminates the measurements varies with each measurement device and is estimated by the GP through the standard deviation $\sigma_a$ in equation 9, for each dataset a. In that manner, the developed framework is able to differentiate high and low-quality measurements and prioritize high-fidelity datasets. In addition, inclinometers were used at both support locations to measure the structural rotation, and a strain gauge was installed at mid-span, on the bottom side of the beam.
The SNRs calculated for the datasets are all higher than 15, and based on Figure~\ref{fig:fig5}, the number of data points provided to the GP algorithm can be defined. In this model, a total of $N_{dp}=7$ is used for all of the available measurement sets, as shown in Figure~\ref{fig:fig12}. The results indicate that the original calculated stiffness value $EI_\mathrm{init}$ is smaller than the real structural stiffness, as the structural responses, in the form of the measurements, tend to be smaller than predicted. The GP framework is able to correctly identify a model that explains the heterogeneous datasets composed of deflections, rotations and strains. In addition, it also ignores the deflection information at approximately $x=0.6L$, as it appears to originate from a malfunctioning sensor. The uncertainty levels, similar to the numerical example, reflect the boundary conditions informed by the GP model in the form of noise-less datasets.

\begin{figure}[h]
  \centering
  \includegraphics{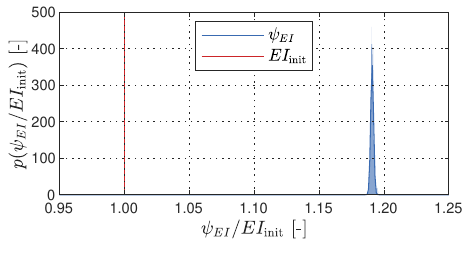}
  \caption{Updated probabilistic bending stiffness model and the initial assumption $EI_\mathrm{init}$.}
  \label{fig:fig13}
\end{figure}

To quantify the updated stiffness value, the probabilistic model of $\psi_{EI}$ is shown in Figure~\ref{fig:fig13}. The normalized $EI_\mathrm{init}$ is used as an initial point for the optimization algorithm, and a stable distribution is obtained with a mean $\mu_{EI}$ that is 19.08\% larger than initially assumed. The model uncertainty, measured in the form of the standard deviation of $\psi_{EI}$, amounts to $\sigma_{EI} = 0.11$\%. 

 \section{Conclusions}
 
A physics-informed Gaussian process model has been developed on the basis of the Euler-Bernoulli beam theory. The model is of a hybrid nature, being driven simultaneously by the heterogeneous data provided to it, as well as the mathematical theory in the form of differential equations. The probabilistic model selection involves the optimization of the GP parameters, the identification of noise in the datasets and, most importantly, the regression of the bending stiffness value in the form of a probability distribution. The identified model can further be used to make probabilistic inferences in any physical quantity linked through the differential equation, even when no data from them is directly provided.
The identified stiffness distribution is directly related to the noise levels of the data provided to the model. A highly noisy dataset requires, therefore, more information to provide a reasonable stiffness prediction. In consequence, a trade-off exists between the data quality and the number of data points provided to the GP model, which leads to higher computation costs. The regressed stiffness parameter, once correctly identified, can be used in a damage identification scenario to reason about the location and the severity of such a damage case.
Finally, an experimental case has been used to showcase that the developed physics-informed GP in a model updating framework. For that matter, the model is also able to integrate heterogeneous datasets, with different measured physical quantities, and different sets of the same quantity with varying levels of noise.

\bibliographystyle{unsrt}  
\bibliography{references}

\end{document}